\documentclass{article}

\usepackage{spconf,amsmath,graphicx}
\usepackage{bm}
\usepackage{amsfonts}

\usepackage{booktabs}   
\usepackage{subfigure}
\usepackage{algorithm}
\usepackage{algorithmic}
\usepackage{lineno,hyperref}
\usepackage{multirow}
\usepackage{tabularx}
\usepackage{pifont}
\usepackage{verbatim}
\usepackage{longtable}

\usepackage[nameinlink]{cleveref}
\Crefformat{equation}{#2Eq.~(#1)#3}

\title{MULTI-TRIAL NEURAL ARCHITECTURE SEARCH WITH Lottery TicketS}

\def\ourmethod{MENAS}
\name{Zimian Wei$^{1}$,  Hengyue Pan$^{1}$, Lujun Li$^{2}$, Menglong Lu$^{1}$, Xin Niu$^{1}$, Peijie Dong$^{1}$, Dongsheng Li$^{1}$}
\address{$^1$ College of Computer, National University of Defense Technology\\
 $^2$ Chinese Academy of Sciences, Beijing, China\\
 }

\begin{document}
%
\maketitle
\begin{abstract}

Neural architecture search (NAS) has brought significant progress in recent image recognition tasks.
Most existing NAS methods apply restricted search spaces, which limits the upper-bound performance of searched models.
To address this issue, we propose a new search space named MobileNet3-MT.
By reducing human-prior knowledge in omni dimensions of networks, MobileNet3-MT accommodates more potential candidates.
For searching in this challenging search space, we present an efficient Multi-trial Evolution-based NAS method termed \ourmethod{}.
Specifically, we accelerate the evolutionary search process by gradually pruning models in the population.
Each model is trained with an early stop and replaced by its Lottery Tickets (the explored optimal pruned network). 
In this way, the full training pipeline of cumbersome networks is prevented and more efficient networks are automatically generated. 
Extensive experimental results on ImageNet-1K, CIFAR-10, and CIFAR-100 demonstrate that \ourmethod{} achieves state-of-the-art performance.


\end{abstract}

\section{Introduction}

The remarkable success of image recognition tasks is largely due to the outstanding neural network architectures. 
Along with this success, many researchers have delivered efforts in designing excellent neural networks,  representative works including ResNet \cite{wightman2021resnet}, MobileNet \cite{sandler2018mobilenetv2}, etc. 
However, manual designing a powerful architecture is tedious and requires a lot of human endeavors. 
Neural architecture search (NAS), which emerged to automate the neural architecture design process, has drawn growing interest and achieved superior performance than human-designed models. 

Different solutions to tackle the NAS problem have been explored in the past few years, as mentioned in \cite{dong2021nats}, they can be coarsely divided into two categories: 
multi-trial NAS \cite{Amobanet, tan2019mnasnet, Enzo2019Neural},
and weight-sharing NAS \cite{cai2019oncecode, liu2018darts, cai2018proxylessnas}.
Multi-trial NAS performs independent training for different networks.  
Existing multi-trial NAS works explore the search space by evolutionary  \cite{Amobanet}, or reinforcement algorithm \cite{tan2019mnasnet}, then train each network from scratch to obtain their performance. 
Most of these methods suffer from high computation costs. 
Weight-sharing NAS trains an over-parameterized supernet, and selects potential sub-net by sharing weights from the supernet \cite{cai2019oncecode, guo2020single}.
Although they have shown appealing efficiency, weight-sharing NAS methods suffer from restricted search spaces, to constrain the supernet for fast convergence \cite{guo2020single, cai2018proxylessnas}, or accommodate all sub-nets on homogeneous building blocks \cite{cai2019oncecode}. 
Limited search space may hinder their practical applications and lead to sub-optimal results due to the human thinking paradigm. 
Comparatively, multi-trial NAS enjoys more flexible application scenarios, and can be applied to more challenging search spaces for better performance.
However, recent literature has been less investigated multi-trial NAS due to the high computation cost.

\begin{figure}[t]
	\begin{center}
		\centerline{\includegraphics[width=0.92\columnwidth]{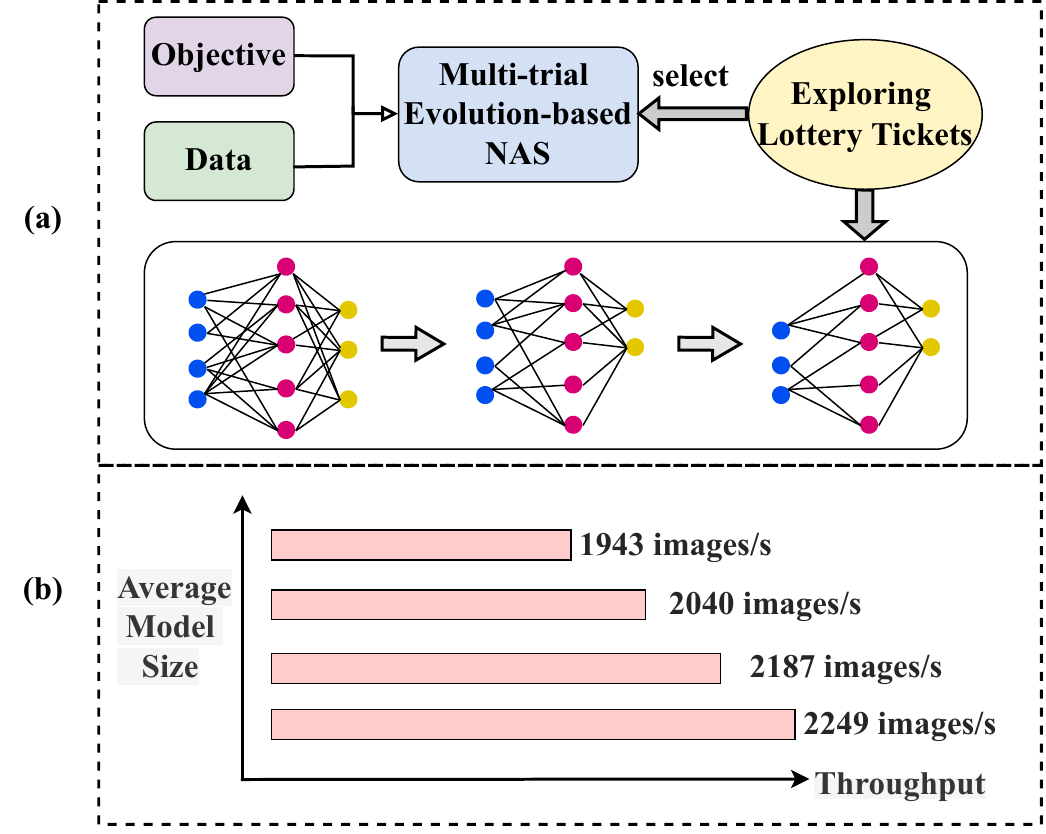}}
		\vspace{-4mm}
		\caption{
		(a) The overview of \ourmethod{}.
       	By jointly exploring Lottery Tickets during the evolutionary search process, the parameters of generated models are gradually reduced.
       	(b) The reduced average model size results in faster training speed, which in turn accelerates the search process.
		}
		\label{fig:overview}
	\end{center}
	\vspace{-10mm}
\end{figure}

\begin{figure*}[t]
	\begin{center}
		\centerline{\includegraphics[width=0.95\textwidth]{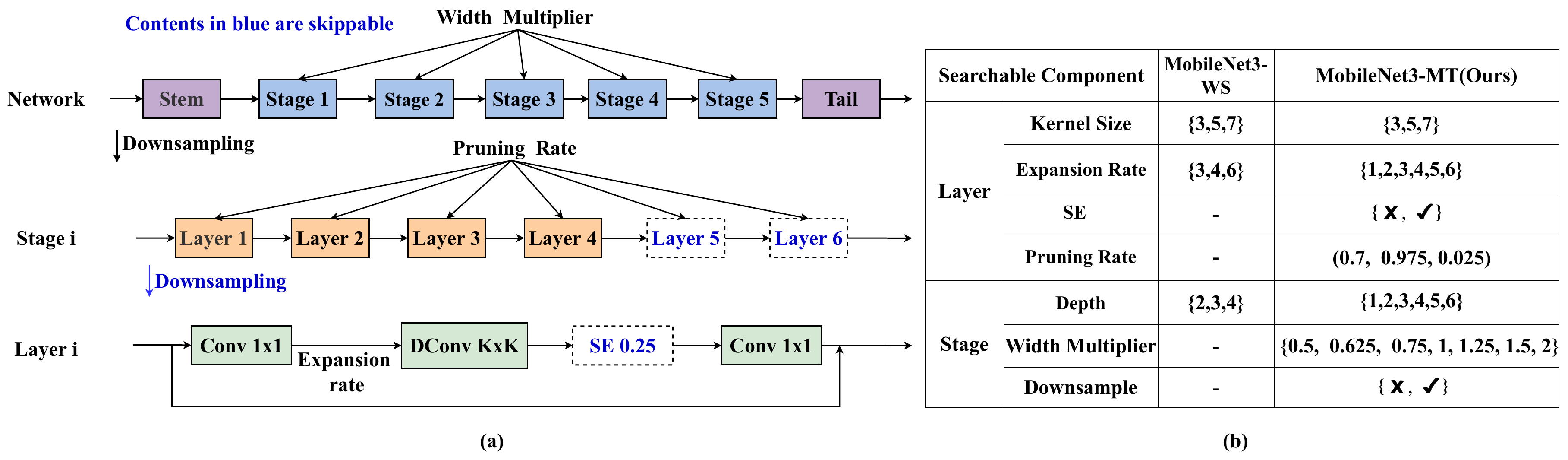}}
		\vspace{-4mm}
		\caption{
			The framework of candidate networks (a) and comparison of searchable components (b) in MobileNet3-MT and existing MobileNet3-WS \cite{cai2019oncecode} search spaces. 
			Each candidate network is grouped into five stages.
			We initialize the output channels of five stages as 16, 32, 64, 128, and 256.
			Each stage contains a variable number of layers, where only the first layer has stride 2 if a down-sample is conducted. 
            Conv is depth-wise convolution in an inverted residual bottleneck, while SE refers to Squeeze-and-Excite module \cite{hu2018squeeze}.
		}
		\label{fig:Mobile_search_space}
	\end{center}
	\vspace{-10mm}
\end{figure*}

In this paper, we focus on multi-trial NAS with a larger search space.
Notably, we draw inspirations from structure pruning works\cite{EagleEye, liu2018rethinking} that provide substantial speedup by removing part of model sub-structures, e.g., channels.
As shown in Fig. \ref{fig:overview}(b), pruned networks enjoy faster training speed. 
The main difference lies in that \ourmethod{} targets to accelerate the search process in NAS and automatically explore optimal pruning strategies (Lottery Tickets) for enormous networks in the search space, while existing pruning methods only build upon several fixed human-designed networks.
We present the overview of \ourmethod{} in Fig. \ref{fig:overview} (a), in which multi-trial evolution-based NAS and exploring Lottery Tickets are conducted jointly.
During the search process, each network is trained with an early stop, then an optimal pruned network (Lottery Ticket) can be searched based on validation accuracy, which has the potential to achieve a similar performance compared with its unpruned counterparts.
The new populations are formulated based on the group of optimal pruned networks from the last generation.
In this way, the average number of model parameters in each generation is gradually reduced and more efficient networks are automatically generated.
Note that the computation cost of exploring Lottery Tickets is negligible since back-propagation is not included.   

Generally speaking, the contributions of this paper are as follows:
1) We propose a multi-trial evolution-based NAS framework with an enlarged search space, in which less prior knowledge is included.  
2) We accelerate the search process by exploring Lottery Tickets, which gradually reduces the average model size in each generation.
3) Extensive experiments on ImageNet-1K, CIFAR-10, and CIFAR-100 demonstrate that \ourmethod{} achieves state-of-the-art performance.

\section{Approach}
We first present the designed search space, then show the details of exploring Lottery Tickets, and finally, we elaborate on the overall evolutionary search pipeline of \ourmethod{}. 

\subsection{Search Space}

We depict our proposed MobileNet3-MT in Fig. \ref{fig:Mobile_search_space}.
Fig. \ref{fig:Mobile_search_space} (a) illustrates the framework of candidate networks, in which the stem and tail are common, while five stages are searchable.
Each stage comprises multiple layers, and each layer is an inverted residual bottleneck.
Comparing with existing search space in recent NAS methods (see MobileNet3-WS in Fig. \ref{fig:Mobile_search_space} (b)), our MobileNetV3-MT differs in following aspects.
First, choices for expansion ratios are extended from \{3, 4, 6\} to \{1, 2, 3, 4, 5, 6\}, while depth choices in each stage are enlarged from  \{2, 3, 4\} to \{1, 2, 3, 4, 5, 6\}. 
Second, each inverted bottleneck is optionally attached with a Squeeze-and-Excite module (SE) \cite{hu2018squeeze}.
Down-sampling is also selective for five stages with a maximum number of 4 (The first down-sample is conducted in the stem structure).
Finally, the width multiplier and pruning rate are searched to determine the number of channels in the five stages and the expansion ratio in each inverted bottleneck layer.
By enlarging the search space, it is favorable to discover more excellent networks.
In the following section, we show details of exploring Lottery Tickets to improve the search efficiency in the huge search space.
\subsection{Exploring Lottery Tickets}
\label{sec:adaptive_prune}
Lottery Ticket \cite{frankle2018lottery} refers to 
the optimal pruned networks capable of training in isolation to match the accuracy of the original network.
For a trained candidate network, we denote each pruning proposal as $(p_1, p_2, ... p_i)$, where $p_i$ is the pruning rate of the ${i}$-th layer.
Based on the pruning proposal, we build a new pruned network and initialize it by the pre-trained weights.
Generally speaking, there are three scenarios for layers in candidate networks:
(1) For convolution layers, the weight is a four-dimension tensor with shape $\left( C_{out},C_{in},K,K \right)$. $K$ is the kernel size of the convolution layer. 
(2) For fully connected layers, the weight is a two-dimension tensor with shape $\left( C_{out},C_{in} \right) $.
(3) For BatchNorm layers, the weight shape is $\left( C_{out}\right) $.
Therefore, the weight matrix $W^*$ of the ${i}$-th pruned layer is obtained as follows:
\begin{equation}\label{eq:weight}
		\footnotesize
	W^*=\begin{cases}
		W\left[ :C_{out}\times p_i,:C_{in}\times p_{i-1},:,: \right] , \quad W\in \mathbb{R}^{\left( C_{out},C_{in},K,K \right)}\\
		W\left[ :C_{out}\times p_i,:C_{in}\times p_{i-1} \right] ,\qquad \,\,    W\in \mathbb{R}^{\left( C_{out},C_{in} \right)}\\
		W\left[ :C_{out}\times p_i \right] ,   \qquad\qquad \qquad \qquad                        W\in \mathbb{R}^{C_{out}}\,\,                   
	\end{cases}
\end{equation}
where $C_{out}$ and $C_{in}$ are numbers of output and input channels for the ${i}$-th layer, respectively. 
$W$ is the pre-trained weights in the original unpruned ${i}$-th layer.

\begin{algorithm}[t]
\footnotesize
	\caption{\quad \footnotesize Overall Evolutionary Search Pipeline}
	\begin{algorithmic}[1]
		\REQUIRE  Generation number $E$, population size $N$, 
		 the mutation and crossover number $\mathcal{M}$ and $\mathcal{S}$,
		training epochs $E_{t}$, fine-tune epoch $E_{f}$,
		Constraints $\mathcal{C}$.
		\FOR{ $e = 1, \dots, E $}
		\FOR{ $i = 1, \dots, N$ }
		\STATE {\bf Evaluation:} 
		training each individual with early stop for $E_{t}$ epochs and compute accuracy;\
		\STATE {\bf Pruning:} exploring Lottery Ticket as in Section \ref{sec:adaptive_prune}, then fine-tune the Lottery Ticket for $E_{f}$ epochs to obtain its accuracy; 
		\ENDFOR
        \STATE {\bf Selection:} 
        select top-k Lottery Tickets with the best accuracy as parent population $\mathcal{G}_{topK}$;
    \STATE {\bf Crossover:} 
    $\mathcal{G}_{crossover}$ = Crossover($\mathcal{G}_{topK},  \mathcal{S}$), s.t. $\mathcal{C}$;\
\STATE {\bf Mutation:} $\mathcal{G}_{mutation}$ = Mutation($\mathcal{G}_{topK}, \mathcal{M}$), s.t. $\mathcal{C}$;
\STATE $\mathcal{G}_{i}$ = $\mathcal{G}_{mutation}$ + $\mathcal{G}_{crossover}$;\
		\ENDFOR
	\ENSURE $\mathcal{G}_{top1}$.
	\end{algorithmic}
	\label{alg_search}
\end{algorithm}

\begin{figure}[t]
	\centering
	\includegraphics[width=0.98\linewidth]{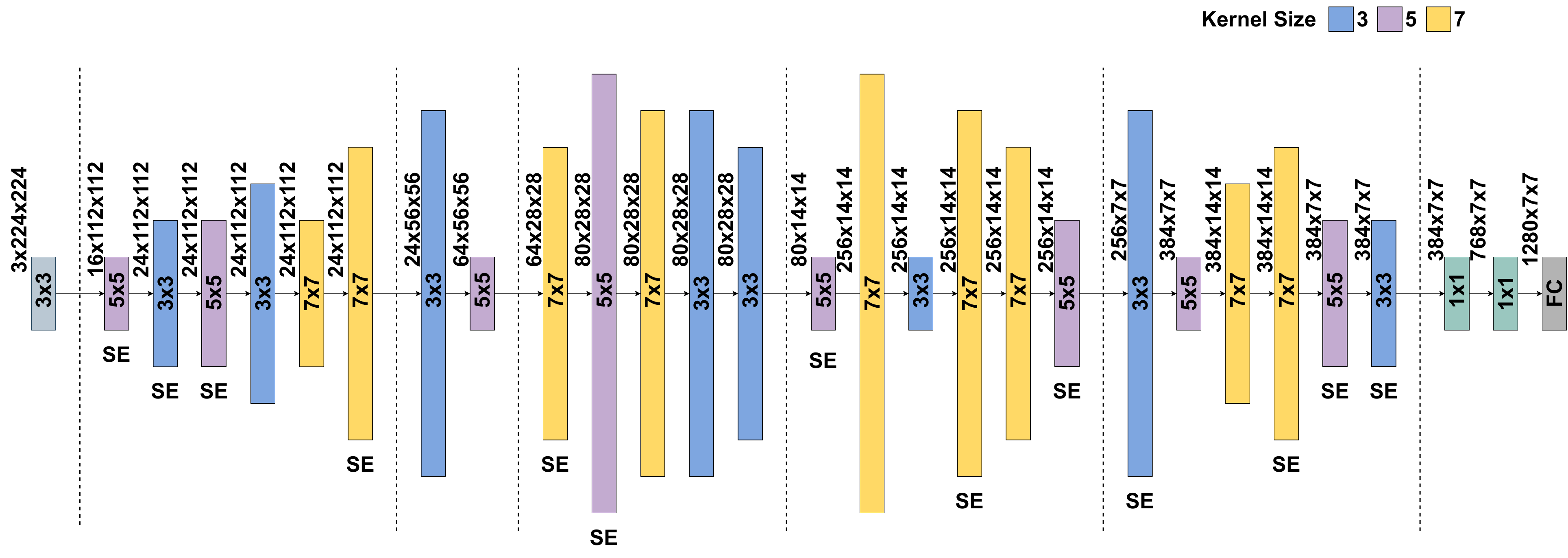}
	\vspace{-4mm}
	\caption{Visualization of searched architecture by \ourmethod{}. 
	SE refers to squeeze-and-excitation \cite{hu2018squeeze} module.
	The height of each layer is proportional to the expansion rate in the inverted residual block.}
	\label{fig:final_arch}
		\vspace{-4mm}
\end{figure}

To rank all pruned networks, an intuitive metric is their validation accuracy.  
Unfortunately, due to the outdated global Batch Normalization statistics that are directly copied from unpruned networks \cite{EagleEye}, all pruned networks show poor and similar validation accuracy.
Therefore, we forward-propagate a few iterations on training data to re-calibrate the statistics in batch normalization. 
Note that when we update the statistics of batch normalization, other trainable parameters of pruned networks are fixed. 
After BatchNorm re-calibration, we can fast evaluate the potential of pruned networks by validation accuracy.
The optimal pruned network is selected as Lottery Ticket to guarantee comparable performance with its unpruned counterpart.

\subsection{Overall Evolutionary Search Pipeline}

We present an overview of our evolutionary search pipeline in Algorithm~\ref{alg_search}.
The parent population is formulated based on best-performing Lottery Tickets.
Crossover and mutation are conducted on parent networks to generate the next generation.
For crossover, two randomly selected parent networks are crossed to produce two new networks. 
For mutation, a randomly selected candidate mutates each element with a probability of 0.08 to produce a new candidate. 
Crossover and mutation are repeated to generate enough new candidates within model computation constraints.
At the end of the evolutionary search process, a final architecture with optimal search accuracy will be returned and then re-trained with the full training schedule.

\begin{table}[t]
		\footnotesize
	\centering
	\caption{Comparison with other state-of-the-art methods on ImageNet-1K.
		'-' denotes manual methods, while 'N/A' means not provided by the origin paper.
		The time costs for \cite{guo2020single, liu2018darts, cai2019oncecode} include supernet training time and search time.
		The time cost for \ourmethod{} includes both model training time and  Lottery Tickets searching time.
		Models marked with $\dagger$ are trained with  knowledge distillation \cite{hinton2015distilling}.
		SPOS 2.2$\times$ and MnasNet-A1 2.3$\times$ are scaled versions with similar model sizes as \ourmethod{} and are trained with the same recipe as ours.
	} 
	\vspace{5pt}
		\resizebox{\columnwidth}{!}{
	\begin{tabular}{lcccc}
		\toprule[1pt]
		Architecture                & FLOPs & Param & Top-1 Acc. & Time Cost      	\\
	 \midrule[0.75pt]
		ResNet-50 \cite{wightman2021resnet}                         & 4.1   & 25.6   & 79.8       & -              \\
		EfficientNet-B2 \cite{tan2019efficientnet}                                            & 1.0   & 9.2    & 80.1       & -              \\ 	\midrule[0.75pt]
		RegNet Y-1.6G \cite{radosavovic2020designing}                   & 1.6   & 11.2   & 78.0       & N/A            \\
		GPUNet-0  \cite{wang2022gpunet}                                 & 3.3  & 11.9   & 78.9       & N/A            \\
		DenseNAS-R3 \cite{fang2020densely}                             & 3.4  & 24.7   & 78.0       & N/A            \\
		RandWire-WS \cite{xie2019exploring}                     & 4.0   & 31.9   & 79.0       & N/A            \\
		SPOS 2.2$\times$~\cite{guo2020single}                                   & 1.5   & 14.5    & 79.3       & 13             \\
		MnasNet-A1 2.3$\times$~\cite{tan2019mnasnet}                         & 1.5   & 13.4    & 79.9       & 288            \\
		OFA-ResNet50D-41 \cite{cai2019oncecode}  & 4.1   & 30.9    & 79.8       & 1600          \\
		\midrule[0.75pt]
		\ourmethod{}                                               & 1.5   & 13.6   & \textbf{80.5}       & 38.4           \\
		\ourmethod{}$\,^{\dagger}$                                        & 1.5      & 13.6        &  \textbf{81.5}          & 38.4               \\
			\bottomrule[1pt]
	\end{tabular}
	}
	\vspace{-6mm}	
	\label{tab:ImageNet-1K}
\end{table}

\section{Experiments}


\subsection{Experimental Setup}
\label{exp_detail}

For the search process, we randomly sample 10\% data of each class from the original 1K-class ImageNet-1K training set for training, while using the original validation set for validation. 
The population is initialized by randomly sampling 60 networks within the model computation budget, which is set as $1\sim 1.6$G FLOPs.
We train our models using the RMSProp optimizer with 0.9 momentum, and a total batch size of 1024 on 8 GPUs. 
The training epoch $E_{t}$, fine-tune epoch $E_{f}$ are set as 30 and 3, respectively.
The learning rate is first warmed up from 0 to 0.064, then decayed by 0.97 every 2.4 epochs. 
For each trained model, we explore Lottery Ticket from randomly generated $50$ pruning proposals.
The total search cost for 5 generations takes 4.8 days on 8 Tesla A100.
We present details of the final architecture in Fig. \ref{fig:final_arch}.
When retraining on the full ImageNet-1K dataset, we use the exponential moving average with 0.9999 decay rate, and RandAugment \cite{cubuk2020randaugment}.
The final architecture is trained from scratch with image size $224 \times 224$ for $300$ epochs. 

\begin{figure*}[th]
\small
\begin{minipage}[t]{0.33\textwidth}
\includegraphics[width=1\textwidth]{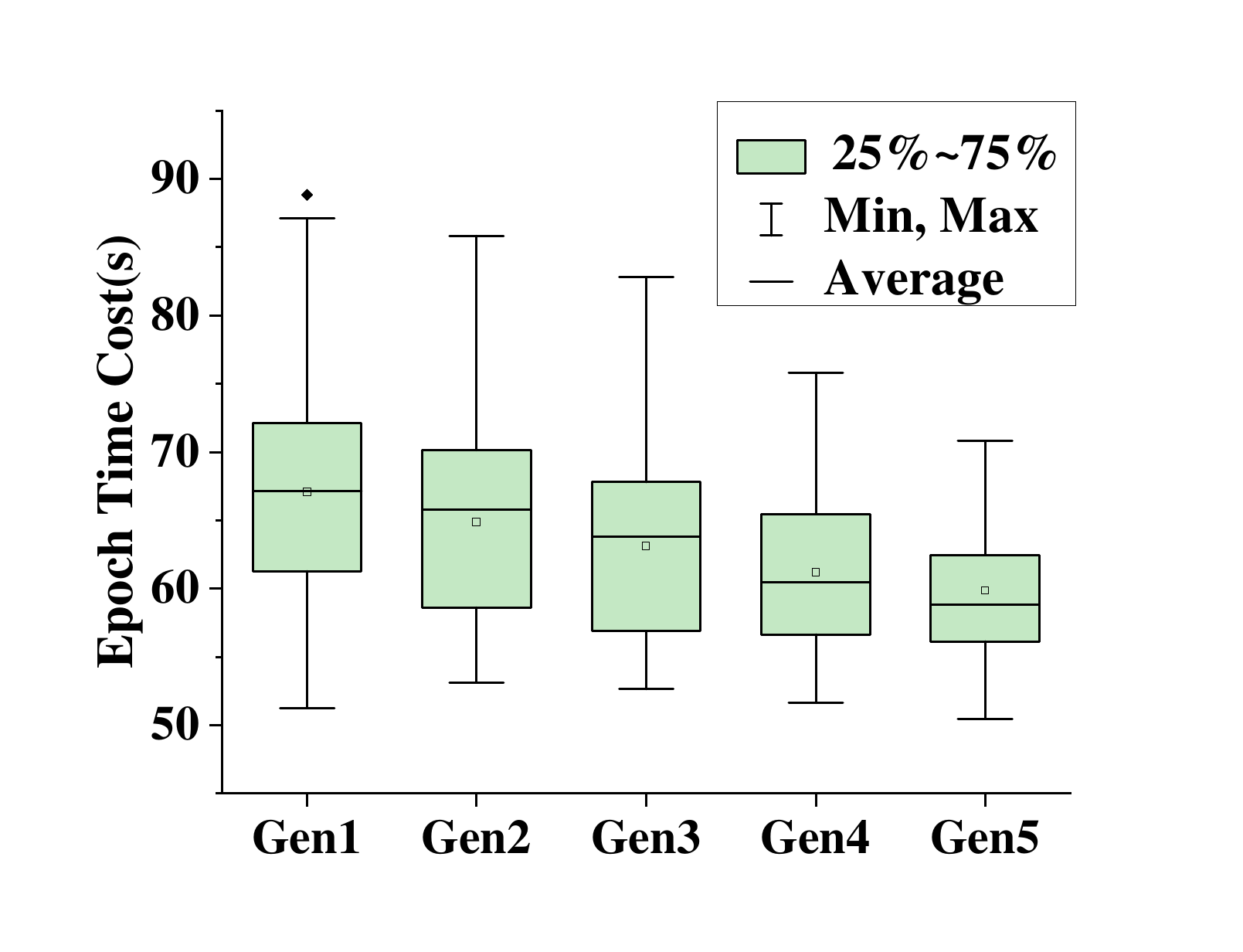}
\centerline{(a)}
\end{minipage}
\begin{minipage}[t]{0.33\textwidth}
\centering
\includegraphics[width=1\textwidth]{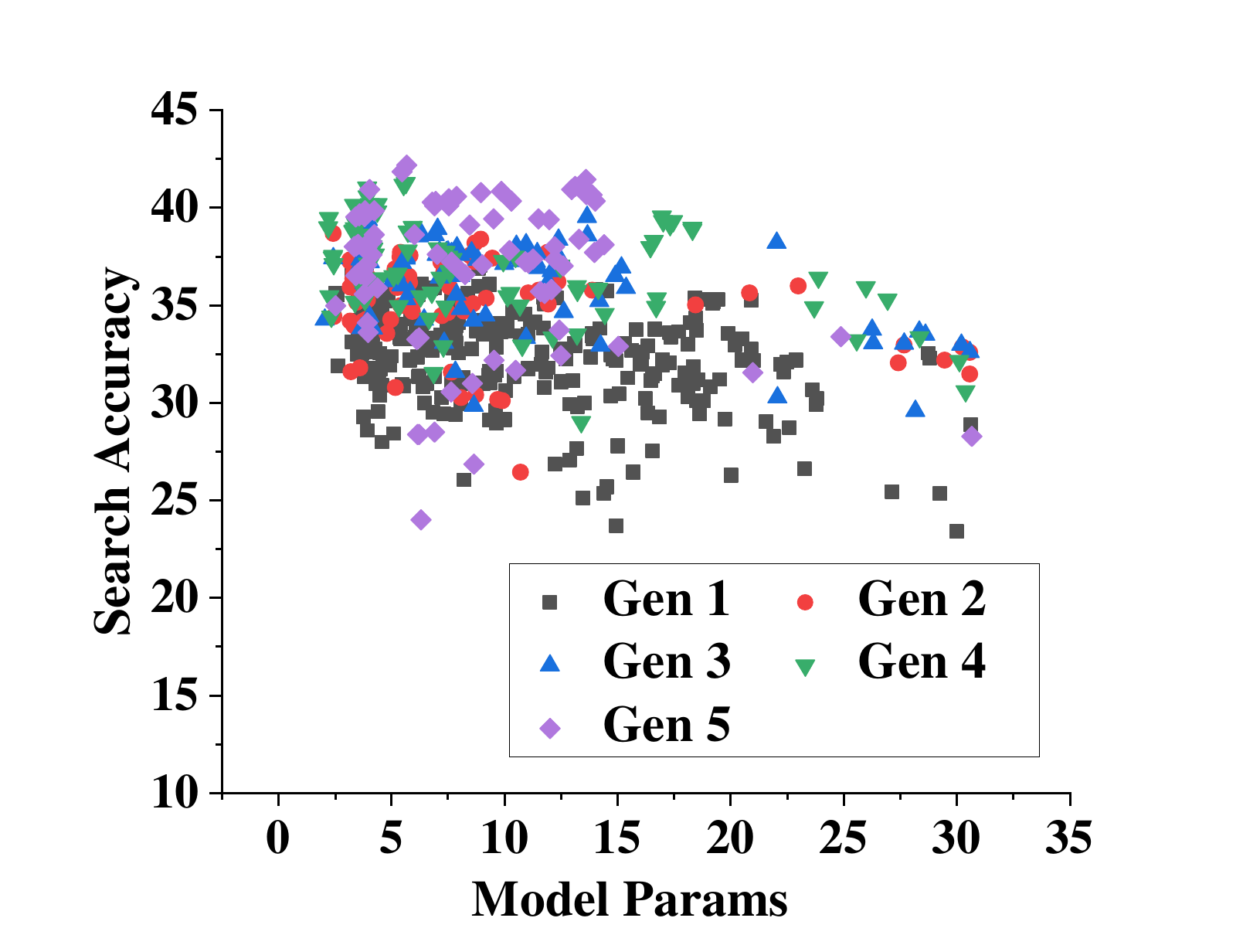}
\centerline{(b)}
\end{minipage}
\begin{minipage}[t]{0.33\textwidth}
\centering
\includegraphics[width=1\textwidth]{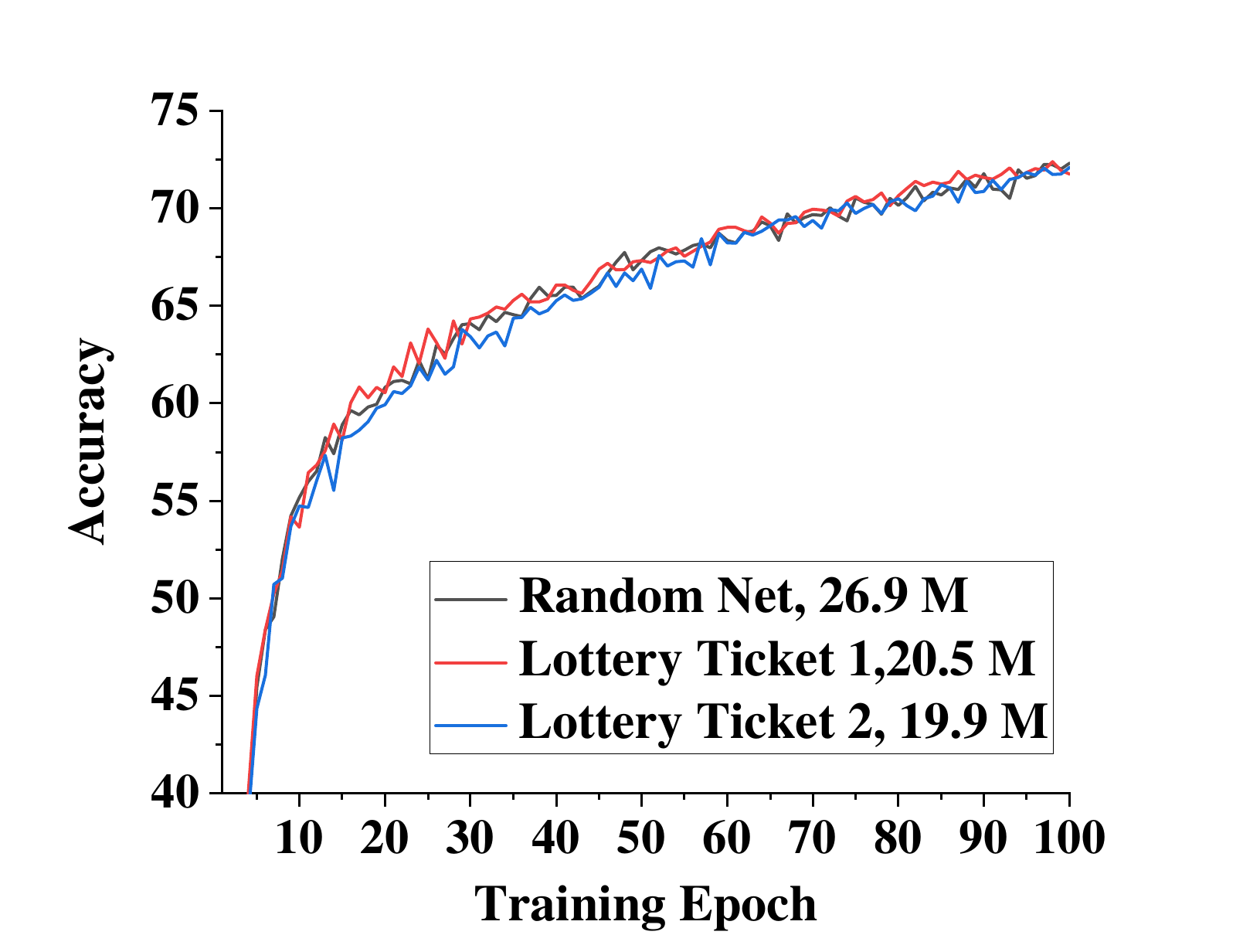}
\centerline{(c)}
\end{minipage}
\caption{
	(a) Distribution of time cost per epoch for networks in each generation. 
	The maximum-smallest range and average line are presented. 
	The time cost for each epoch includes training time, validation time, and data loading time in the search process. 
		(b) Distribution of model parameters and search accuracy of networks in 5 generations during the evolutionary search process.  
		(c) 100-epoch retraining curve on the full ImageNet-1K for a random model and its explored two Lottery Tickets.
}
\label{fig:Lottery_Tickets}
\vspace{-5mm}
\end{figure*}

\subsection{Results on ImageNet-1K}
Table \ref{tab:ImageNet-1K} presents the comparison of \ourmethod{} with other state-of-the-art models. 
Without any bells and whistles, \ourmethod{} achieves 80.5\% top-1 accuracy, surpassing EfficientNet-B2 \cite{tan2019efficientnet} by 0.4\%. 
When comparing with other NAS methods, \ourmethod{} shows a boost of 0.6\% top-1 accuracy on MnasNet-A1 2.3$\times$~\cite{tan2019mnasnet} while requiring significantly less search cost.
Notably, when trained with knowledge distillation \cite{hinton2015distilling}, \ourmethod{} further achieves 81.5\% top-1 accuracy, demonstrating our method's strong capacity.

\begin{table}[t]
	\footnotesize
	\caption{Transfer learning results on 
	downstream classification datasets CIFAR-10 and CIFAR-100. 
	Since label classes have changed, the model parameter is reduced to 12.5 M. 
	'N/A' means not provided by the origin paper.
	}
	\centering
	\vspace{5pt}
		\begin{tabular}{lcccc}
			\toprule[1pt]
			Model & Param    & CIFAR-10 & CIFAR-100  \\
			\midrule[0.75pt]
			$\text{ViT-B/16}$ \cite{dosovitskiy2020image}& 87M   & 98.1 & 87.1  \\
			$\text{NASNet-A}$ \cite{Enzo2019Neural}& 85M  & 98.0 & 87.5    \\ 
			$\text{ResNet50-A1}$ \cite{wightman2021resnet}   & 25.6 M    & 98.3  & 86.9 \\
			$\text{ResMLP-S12}$ \cite{touvron2021resmlp}& 15.4M   & 98.1 & 87.0  \\
			\midrule[0.75pt]
			\ourmethod{} & 12.5M  &  \textbf{98.5} & \textbf{88.3}   \\
			\bottomrule[1pt]
		\end{tabular}
	\label{tab:transfer}
	\vspace{-6mm}
\end{table}

\begin{table}[t]
	\footnotesize
	\caption{Comparison with the random baseline of coarsely identical search time.
	$N$ refers to the number of evaluated networks in the search process.
	SS represents search space.
	For \ourmethod{}, the evaluated networks include normally trained networks and fine-tuned Lottery Tickets. 
		}
			\centering
	\vspace{5pt}
\begin{tabular}{ccccc}
	\toprule[1pt]
Method                         & SS            & N     & Params & Top-1 Acc. \\ 	\midrule[0.75pt]
\multirow{2}{*}{Random Search} & MobileNet3-WS & N=200 &  11.8 M      &  78.6          \\
                               & MobileNet3-MT & N=200 & 16.9 M & 79.4       \\ 	\midrule[0.75pt]
Ours                           & MobileNet3-MT & N=540 & 13.6 M & 80.5       \\ 	\bottomrule[1pt]
\end{tabular}
\label{exp_random}
	\vspace{-6mm}
\end{table}


\subsection{Results on CIFAR-10 and CIFAR-100}

We transfer ImageNet pre-trained \ourmethod{} to downstream datasets CIFAR-10 and CIFAR-100.
Table \ref{tab:transfer} shows the results of top-1 accuracy for transfer learning. 
In general, our method consistently reports better accuracy than other state-of-the-art models, suggesting good generality of \ourmethod{}.

\subsection{Effect of Lottery Ticket Search}

To validate the effectiveness of exploring Lottery Tickets in \ourmethod{}, we present epoch time cost distribution in Fig.\ref{fig:Lottery_Tickets} (a) and search accuracy vs. model size distribution for 5 generation in Fig. \ref{fig:Lottery_Tickets} (b).
Epoch time represents the training speed of networks, which is much related to model parameters and computations.
It can be observed that due to the substantial speedup provided by pruning, the average epoch time cost is gradually reduced during the searching process.
Fig. \ref{fig:Lottery_Tickets} (b) further indicates that the evolutionary process is guided to more efficient but well-performed networks by generations. 
Moreover, we retrain a randomly sampled architecture and its two Lottery Tickets under the same setting to compare their performance. 
The training curves are depicted in Fig.~\ref{fig:Lottery_Tickets} (c).
It is evident that Lottery Tickets show comparable performance with reduced model sizes, supporting our motivations.

\subsection{Comparison with Random Search}

We compare \ourmethod{} with the random baseline of coarsely identical search time.
Each search process is conducted on a 10\% subset of ImageNet-1K, then the selected best network is re-trained on the full dataset. 
The model computation constraint (1$\sim$1.6 G FLOPs) for random baselines are the same as \ourmethod{}.
Results are presented in Table \ref{exp_random}.
It can be observed that \ourmethod{} consistently outperforms the random baselines, indicating the superior of our method.

\section{Conclusion}

In this paper, we propose \ourmethod{}, a multi-trial NAS framework with an enlarged search space for higher performance. 
\ourmethod{} improves search efficiency by exploring Lottery Tickets, which gradually prunes the searched models in each generation and results in substantial speedup for training.
State-of-the-art performance on ImageNet-1K, CIFAR-10, and CIFAR-100 demonstrate the effectiveness of \ourmethod{}.

\clearpage

\bibliographystyle{IEEEbib}
\bibliography{mybibfile}

\end{document}